\documentclass{article}

\usepackage[numbers]{natbib}

\usepackage[preprint]{neurips_2024}

\usepackage[dvipsnames]{xcolor}         
\definecolor{linkColor}{rgb}{0.18,0.39,0.62}
\usepackage{colortbl} 
\usepackage[utf8]{inputenc} 
\usepackage[T1]{fontenc}    
\usepackage[colorlinks=true,linkcolor=linkColor,citecolor=linkColor,filecolor=linkColor,urlcolor=linkColor]{hyperref}       

\usepackage{url}            
\usepackage{booktabs}       
\usepackage{amsfonts}       
\usepackage{nicefrac}       
\usepackage{microtype}      

\usepackage{multirow}
\usepackage{amsmath}
\usepackage{amssymb}
\usepackage{graphicx}
\usepackage{pifont}
\usepackage{textcomp}
\usepackage[most]{tcolorbox}
\usepackage{color}
\usepackage{xcolor}
\usepackage{enumitem} 
\usepackage{wrapfig}
\usepackage{subfigure}
\usepackage{tablefootnote}

\newcommand{\ie}{\textit{i.e.}}

\newcommand{\ours}{\texttt{MoCa}}

\newcommand{\mlmT}{\tilde{T}^\text{MLM}}
\newcommand{\mlmX}{\tilde{x}^\text{MLM}}
\newcommand{\maeT}{\tilde{T}^\text{MAE}}

\newcommand{\ourX}{\tilde{x}^\text{\ours}}
\newcommand{\mlmLoss}{\mathcal{L}_\text{MLM}}
\newcommand{\maeLoss}{\mathcal{L}_\text{MAE}}
\newcommand{\ourLoss}{\mathcal{L}_\ours}

\newcommand{\fcausal}{f_{\theta}^{\text{causal}}}
\newcommand{\Embcausal}{\text{Emb}_{\theta}^{\text{causal}}}
\newcommand{\fbi}{f_{\theta}^{\text{bi}}}
\newcommand{\Embbi}{\text{Emb}_{\theta}^{\text{bi}}}

\title{\ours{}: Modality-aware Continual Pre-training Makes Better Bidirectional Multimodal Embeddings}

\author{Haonan Chen$^{1}$\thanks{Work done during Haonan’s internship at Microsoft Research Asia.}, Hong Liu$^{2}$, Yuping Luo, Liang Wang$^3$, \\ \textbf{Nan Yang$^3$, Furu Wei$^3$, Zhicheng Dou$^{1}$} \\
         $^1$Gaoling School of Artificial Intelligence, Renmin University of China \\ 
         $^2$Stanford University $^3$Microsoft Corporation  \\ 
         \texttt{\{hnchen,dou\}@ruc.edu.cn} \\
        \texttt{hliu99@cs.stanford.edu, yupingl@cs.princeton.edu} \\
         \texttt{\{wangliang,nanya,fuwei\}@microsoft.com} \\
         \url{https://haon-chen.github.io/MoCa/} \\
}

\newtcolorbox[list inside=prompt]{prompt}[1][]{
    colbacktitle=black!60,
    coltitle=white,
    fontupper=\footnotesize,
    boxsep=5pt,
    left=0pt,
    right=-1pt,
    top=0pt,
    bottom=0pt,
    boxrule=1pt,
    width=\textwidth,
    #1,
}
\begin{document}
\maketitle

\begin{abstract}

Multimodal embedding models, built upon causal Vision Language Models (VLMs), have shown promise in various tasks. However, current approaches face three key limitations: the use of causal attention in VLM backbones is suboptimal for embedding tasks; scalability issues due to reliance on high-quality labeled paired data for contrastive learning; and limited diversity in training objectives and data. To address these issues, we propose \ours{}, a two-stage framework for transforming pre-trained VLMs into effective bidirectional multimodal embedding models. The first stage, Modality-aware Continual Pre-training, introduces a joint reconstruction objective that simultaneously denoises interleaved text and image inputs, enhancing bidirectional context-aware reasoning. 
The second stage, Heterogeneous Contrastive Fine-tuning, leverages diverse, semantically rich multimodal data beyond simple image-caption pairs to enhance generalization and alignment. 
Our method addresses the stated limitations by introducing bidirectional attention through continual pre-training, scaling effectively with massive unlabeled datasets via joint reconstruction objectives, and utilizing diverse multimodal data for enhanced representation robustness.
Experiments demonstrate that \ours{} consistently improves performance across MMEB and ViDoRe-v2 benchmarks, achieving new state-of-the-art results, and exhibits strong scalability with both model size and training data on MMEB.

\end{abstract}
\section{Introduction}

Multimodal embedding models have achieved significant improvements in various tasks including multimodal classification, visual question answering, and document retrieval~\citep{MMEB,GME,mme5,DSE}.
These models are often built on Vision Language Models (VLMs), such as Phi-V~\cite{Phi3}, LLaVA~\citep{Llava}, and Qwen-VL~\citep{Qwen25-VL}, which demonstrate strong generation and comprehension capabilities across modalities.
Recent methods, including VLM2Vec~\citep{MMEB} and mmE5~\citep{mme5}, apply contrastive learning on image-text pairs to off-the-shelf VLMs to align modalities and improve cross-modal representation quality.

Despite the growing success of multimodal embedding models, three main limitations remain in current approaches:
(1) \textbf{Causal attention of pre-trained VLMs might be suboptimal for embedding models.} 
Mainstream multimodal embedding models~\citep{MMEB,mme5,GME} inherit causal attention from their VLM backbones~\citep{llama3,Llava,Phi3,Qwen2-VL}.
However, studies on text embedding models~\citep{LLM2Vec,GTE,NV-Embed,Echo} have shown that bidirectional attention typically produces superior embeddings compared to causal attention. This finding is reinforced by seminal multimodal works such as CLIP~\citep{CLIP} and SigLIP~\citep{SigLIP}, which adopted bidirectional encoders after extensive architectural exploration.
Furthermore, bidirectional embedding models with mean pooling offer practical benefits, such as late chunking~\cite{Late_Chunking}. 
This suggests a strong need to evaluate bidirectional architectures for multimodal embedding models.  
(2) \textbf{Contrastive learning is hard to scale without labeled pair data.}
Contrastive learning fundamentally depends on diverse high-quality image-text pairs, which limits its scalability. Although large datasets of image-caption pairs exist~\citep{LAION,SigLIP2}, curating diverse and high-quality multimodal pairs remains resource-intensive and hard to scale. Moreover, contrastive learning cannot leverage the vast amount of unpaired multimodal data available on the internet. These limitations underscore the need for more scalable training paradigms that can effectively utilize diverse, unlabeled multimodal data.
(3) \textbf{Lack of diversity in training objectives and data distribution leads to suboptimal cross-modal alignment.} Prior works such as~\citet{MMEB,mme5} typically fine-tune Vision-Language Models (VLMs) with a single contrastive objective applied to a narrow range of data, \ie, mostly short image-caption pairs. This setup fails to fully exploit the rich cross-modal reasoning capabilities that pre-trained VLMs offer.
While such training yields aligned embeddings, it encourages alignment based primarily on surface-level similarity, rather than deeper semantic or contextual understanding. As a result, the learned embedding models often overfit to the training distribution and struggle to generalize to more complex or diverse multimodal scenarios. 
This suggests the need for more expressive training objectives that explicitly promote cross-modal reasoning over heterogeneous data.

Recent advances in text embedding models, such as LLM2Vec~\citep{LLM2Vec}, adapt pre-trained language models using bidirectional Masked Language Modeling (MLM)~\citep{bert}. Despite the success of Continual Pre-Training (CPT) in text-only domains, its potential remains underexplored for multimodal embeddings. Moreover, as shown in Section~\ref{sec:ablation}, MLM alone is insufficient for handling mixed-modality inputs, motivating the need for modality-aware, bidirectional objectives that can jointly model interleaved image and text signals.

\begin{figure*}[t]
  \centering 
   \subfigure{
    \includegraphics[width=0.30\textwidth]{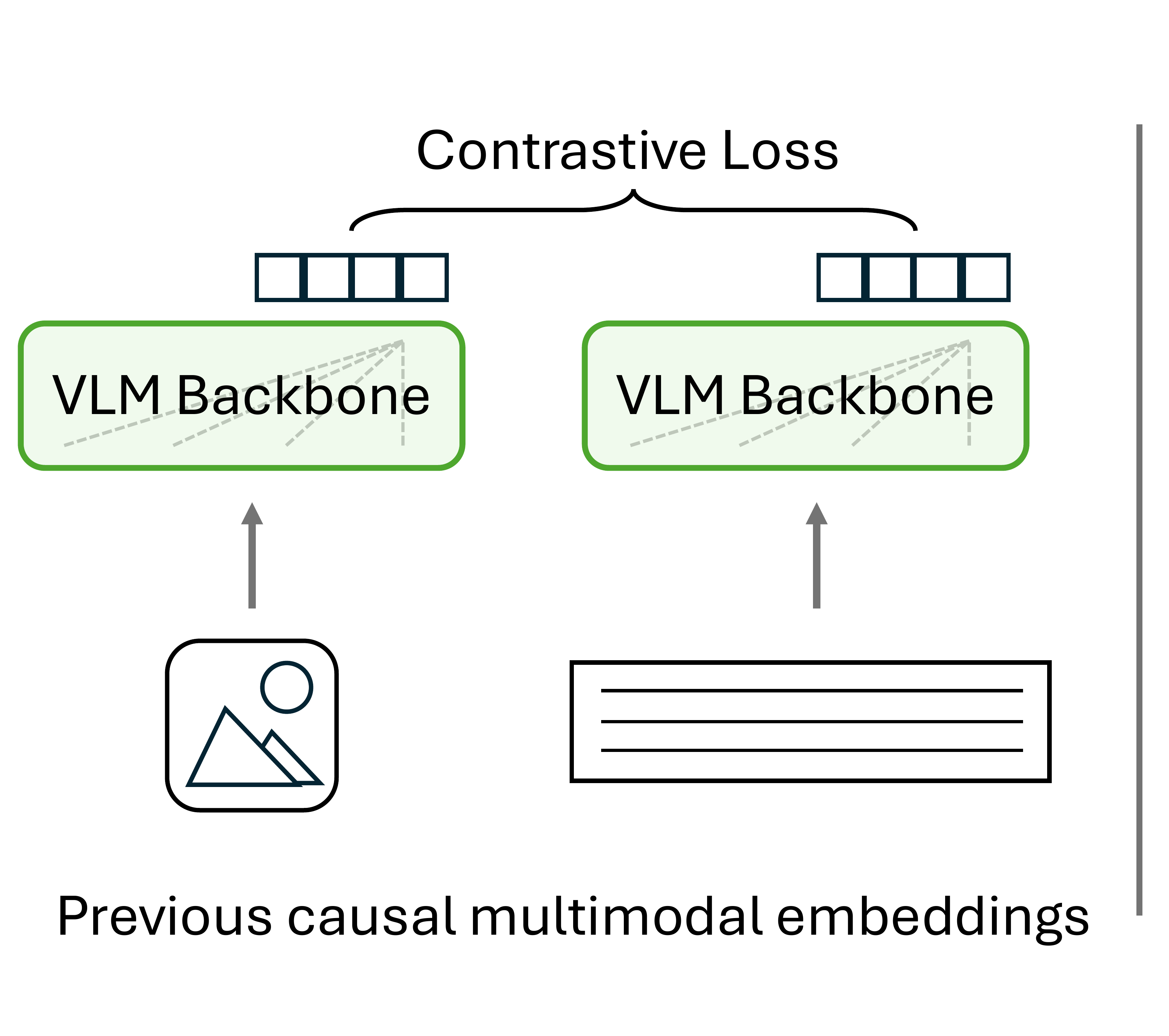} 
    }
    \hspace{1pt}
   \subfigure{
    \includegraphics[width=0.618\textwidth]{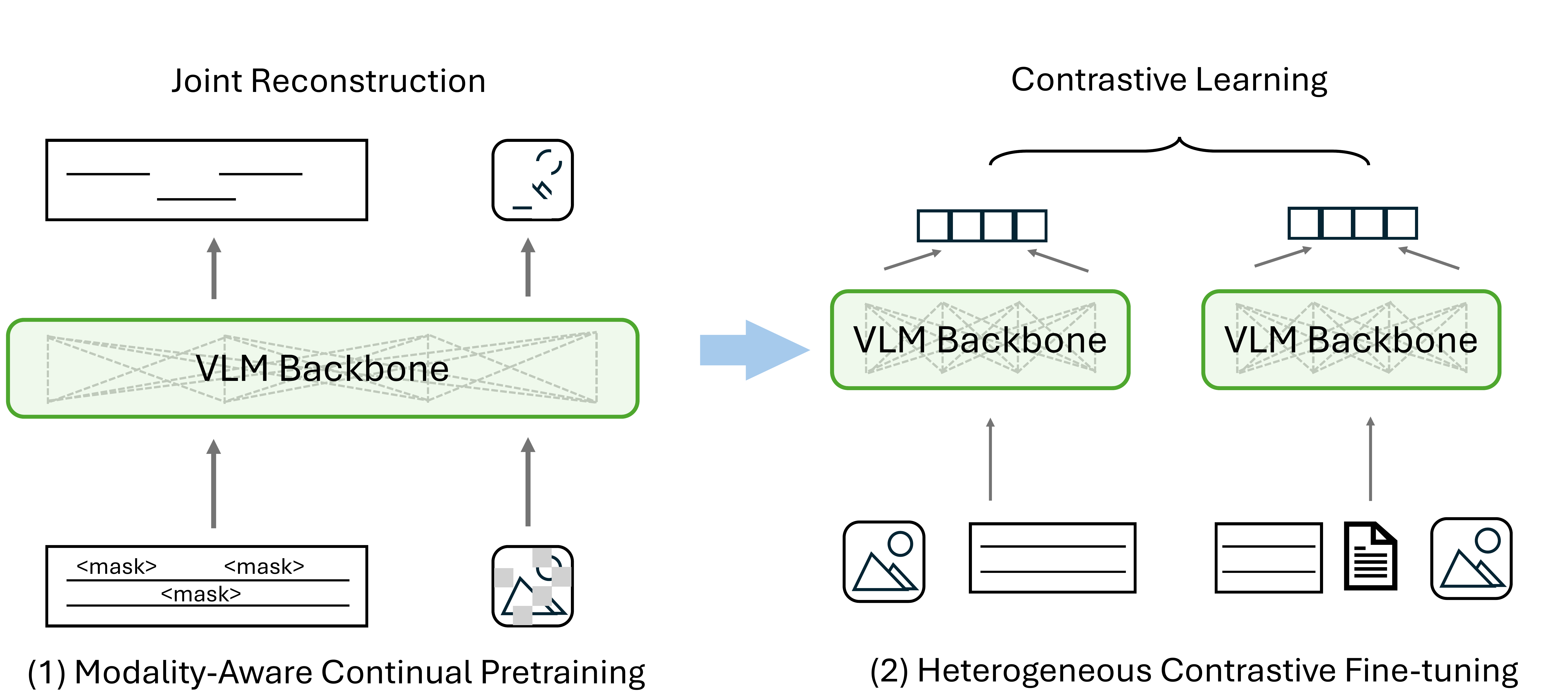}
    }
    \vspace{-5pt}
  \caption{\textbf{Comparison of VLM-based multimodal embedding models.} Left: Previous single-stage contrastive learning with mainly image-caption pairs and causal attention. Right:~\ours{}. In \textit{modality-aware continual pre-training}, we optimize a joint bidirectional reconstruction objective to denoise interleaved texts and images simultaneously. In \textit{heterogeneous contrastive fine-tuning}, we train models to improve cross-modal fusion of the backbone with diverse image and text contexts.} 
  \label{fig:main}
  \vspace{-10pt}
\end{figure*}

In this work, we introduce a two-stage framework, \ours{}, to transform pre-trained VLMs into effective bidirectional multimodal embedding models. As illustrated in Figure~\ref{fig:main}, our approach consists of two stages: (1) \textbf{Modality-aware Continual Pre-training} and (2) \textbf{Heterogeneous Contrastive Fine-tuning}. 
In the first stage, we introduce a joint reconstruction objective that requires the model to simultaneously denoise interleaved text and image inputs, which encourages the model to jointly reason across modalities. 
For text, we apply masked language modeling (MLM), where masked tokens are predicted using the full multimodal context. 
For images, we adopt masked autoencoding (MAE): a subset of image patches are randomly masked and reconstructed by a lightweight decoder conditioned on image and text contexts. 
In the second stage, as opposed to previous works which used mainly image-caption pairs, we add diverse heterogeneous data including (i) long-form query-document pairs, supporting document-level understanding and complex reasoning over extended context, (ii) curated multimodal pairs, offering various visual and textual contexts beyond image captions of specific distributions, and (iii) real-world text pairs, enhancing linguistic representations across diverse domains.

Together, the two stages directly address the limitations outlined above.
Stage one tackles Limitations (1) and (2) by using massive unlabeled interleaved data and enhancing bidirectional, context-aware reasoning across modalities.
It also partially mitigates Limitation (3) by applying joint reconstruction on diverse multimodal inputs.
Stage two directly addresses Limitation (3) by introducing heterogeneous and semantically rich multimodal pairs to enhance generalization and alignment across various domains.

We conduct experiments with~\ours{} and verify that
our trained model consistently improves performance across MMEB~\citep{MMEB} and ViDoRe-v2~\citep{ColPali} benchmarks.
Besides, it demonstrates strong scalability with respect to both model size and training data on MMEB.
Specifically, after continual pre-training on only 30B tokens, our 3B model matches or surpasses the performance of competitive 7B baselines. 
When scaled to 7B parameters, our model sets new state-of-the-art results on MMEB.
Furthermore, the performance improves steadily as the corpus for CPT grows, indicating the framework’s effectiveness in leveraging more and broader unlabeled multimodal data.

In summary, our contributions are as follows.

\begin{itemize}[leftmargin=2.5mm]

\item We are the first to propose a continual pre-training (CPT) approach with unlabeled data to adapt VLMs to bidirectional embedding models and demonstrate its strong scalability with respect to model and corpus sizes.

\item We also show that contrastive fine-tuning with heterogeneous data and cross-modal interactions enhances model generalization.

\item With the two techniques combined, our framework, \ours{}, consistently improves performance on various benchmarks and achieves state-of-the-art performance on MMEB.

\end{itemize}

\section{Method: \ours{}}

In this section, we present~\ours{}, which transforms pre-trained vision language models (VLMs) into powerful bidirectional multimodal embedding models. 
As illustrated in Figure~\ref{fig:main2}, our method comprises two stages:
(1)~\textbf{Modality-aware Continual Pre-training}, where the model learns to reconstruct masked textual and visual inputs with a joint denoising objective. The reconstruction objective leverages masked language modeling (MLM) and masked autoencoding (MAE), which enables the model to shift from causal to bidirectional attention with better representation quality. Moreover, learning to jointly reconstruct images from text
and text from images enhances cross-modal alignment and reasoning.
(2)~\textbf{Heterogeneous Contrastive Fine-tuning}, where we perform contrastive fine-tuning with a diverse set of multimodal pairs spanning long-form query-document pairs, curated multimodal pairs and real-world text pairs.
This stage further aligns vision and language embeddings while improving generalization across varied real-world tasks. Together, the two stages enable VLMs to produce robust representations suitable for a wide range of downstream applications.

\subsection{Preliminaries}

\paragraph{Vision Language Model Backbones.} Vision language models (VLMs) are widely adopted as the backbones of multimodal embedding models~\citep{MMEB,GME,mme5}. Consider a multimodal input of length $T$, denoted by $x = [x_1, \dots, x_T]$, where each $x_{i}, i \in [T]$, can be either a discrete text token or a continuous image patch. A VLM backbone first maps the input to the same input embedding space with input embedding layers for text tokens and visual encoders for image patches. 
The input embeddings are then passed through the backbone transformer to obtain a series of hidden states $\fcausal(x) \in \mathbb{R}^{T \times d}$ where
$\theta$ denotes the model parameters, and $\fcausal(\cdot)$ refers to the VLM backbone with causal attention. 

\paragraph{Multimodal embedding models.} \citet{MMEB,GME,mme5} inherit the causal attention from VLM backbones. Therefore, $(\fcausal(x))_j$ only depends on $[x_1, \dots, x_j]$. To extract embeddings from the backbone, a natural choice would be the hidden states corresponding to the last (EOS) token, i.e. $\Embcausal(x) := {(\fcausal(x))}_T$. 

Despite its practical success, text embedding studies~\citep{LLM2Vec,GTE,NV-Embed,Echo} found that causal embedding models are worse than bidirectional ones in terms of representation quality. To this end, we introduce bidirectional VLM backbone $\fbi(\cdot)$, which essentially removes the attention causal masks from $\fcausal(\cdot)$. To extract embeddings from $\fbi(\cdot)$, we adopt mean pooling~\citep{reimers2019sentence}. Therefore we have $\Embbi(x) := \frac{1}{T}\sum\limits_{i=1}^T{(\fbi(x))}_i$.

\paragraph{Contrastive learning.} Based on the causal multimodal embedding models initialized with pre-trained VLMs above, previous works follow the recipe of text embedding models~\citep{E5,GTE,NV-Embed,bge} to align image and text embeddings. Each training example is a tuple $(q, d^+, \{d^-_1, \dots, d^-_K\})$, where $q$ is the \textit{query}, $d^+$ is a \textit{positive document}, and $\{d^-_1, \dots, d^-_K\}$ is a set of \textit{hard negative documents}.

\subsection{Modality-aware Continual Pre-training}

\begin{figure*}[t]
  \centering 
    \includegraphics[width=0.95\textwidth]{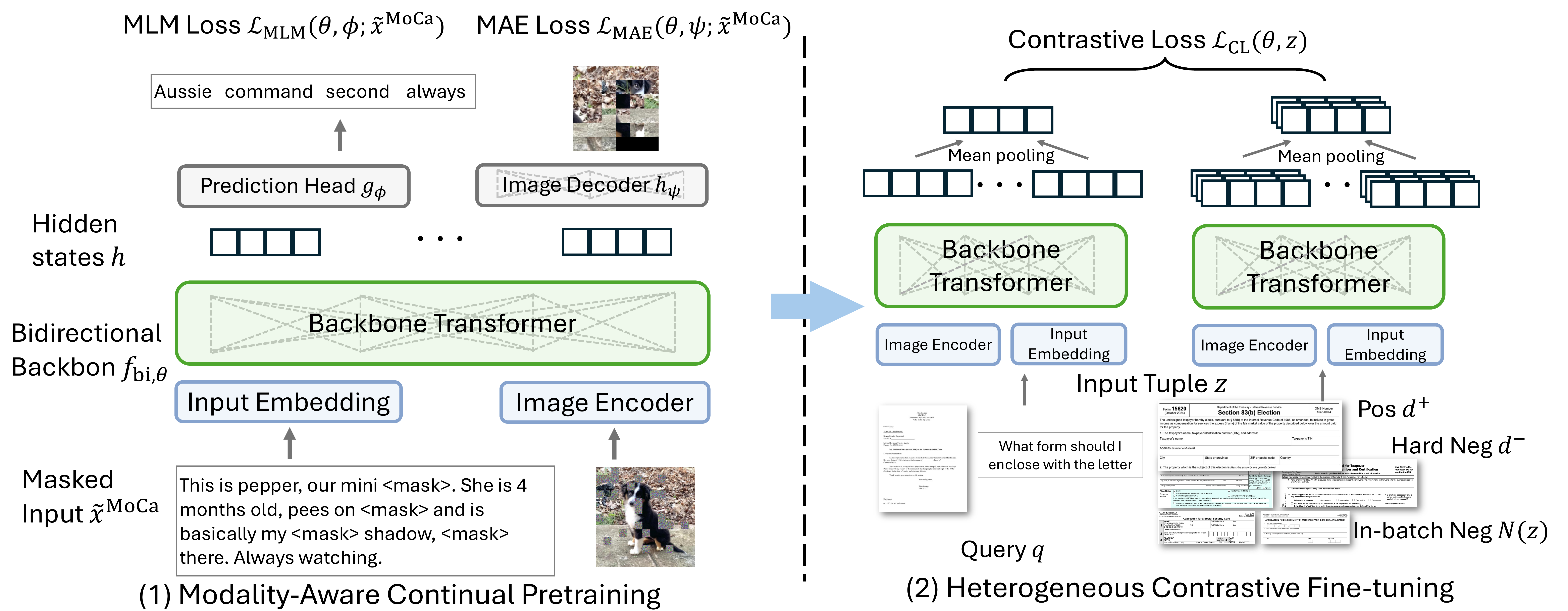} 
    \vspace{-5pt}
  \caption{\textbf{\ours.} (1) In \textit{modality-aware continual pre-training}, the VLM backbone is trained to jointly reconstruct masked texts and images based on interleaved multimodal context with masked language modeling and masked autoencoding, respectively. (2) In \textit{heterogeneous contrastive fine-tuning}, the VLM backbone from the previous stage is further fine-tuned with contrastive loss on a broad range of heterogeneous data. For each query, we curate positive documents, hard negative documents. We apply task batching to make sure in-batch negatives come from the same task. } 
  \label{fig:main2}
  \vspace{-10pt}
\end{figure*}

This stage aims to enhance the bidirectional representation capabilities of a pre-trained VLM by a joint denoising objective over both textual and visual inputs. 
As shown in the left part of Figure~\ref{fig:main2}, we incorporate two complementary objectives: masked language modeling (MLM)~\citep{bert} for texts and masked autoencoding (MAE)~\citep{MAE} for images.

\paragraph{Masked Language Modeling.}

We apply MLM to the text tokens in the input $x$. 
Specifically, we randomly sample a subset of text tokens $\mlmT \subset \{1, \dots, T\}$ and replace each $x_i, \ i \in \mlmT$, with a special mask token \texttt{<|mask|>}. Note that compared to the original input $x$, only selected text tokens $x_i, \ i \in \mlmT$, are changed to \texttt{<|mask|>}. Other text tokens and image patches are left unchanged in text masking. The resulting input with masked text tokens is denoted as $\mlmX$.

The VLM encoder then processes the sequence with masked text tokens with bidirectional attention, enabling each masked position to attend to all visible text tokens and image patches. This facilitates contextual learning that captures both intra-modal and cross-modal dependencies. 
The model is trained to accurately predict each masked token $x_i, \ i \in \mlmT$, based on the surrounding context.
Following~\citet{LLM2Vec}, we predict the masked token at position $i$ using the representation of the previous token ${i-1}$ (\ie, shift the labels) to align with the training recipe of most auto-regressive VLMs.
Suppose the MLM prediction head is $g_{\phi}(\cdot)$ parameterized by $\phi$. The MLM loss is then computed with cross-entropy over the masked positions:
\begin{equation}
\mlmLoss(\theta, \phi; \mlmX) = \sum_{i \in \mlmT} \ell_{\textup{CE}}((p_{\theta,\phi}(\mlmX))_{i-1}, x_i),
\end{equation}
where $(p_{\theta,\phi}(\mlmX))_{i-1}:=g_\phi({(f_{\text{bi},\theta}(\mlmX))}_{i-1})$ is the model’s predicted distribution over the vocabulary at position $i$. 
This objective encourages the model to leverage both local and global contexts to recover masked information, enhancing its ability as a bidirectional encoder for embedding tasks.

\paragraph{Masked Autoencoding.}
Similar to MLM on text tokens, we want a denoising objective on image patches to reconstruct image patches based on text and image context. Inspired by~\citet{MAE}, we mask image tokens and feed them to the bidirectional VLM backbone. On top of the VLM hidden states, we use a light-weight decoder to predict the original patches. Concretely, given input with masked text tokens $\mlmX$, we further randomly sample a subset of image patches $\maeT \subset \{1, \dots, T\}$ and replace each $x_i, \ i \in \maeT$, with vectors sampled from a unit Gaussian distribution. The resulting input with masked text and images is denoted as $\ourX$. 

The model is trained to accurately predict each masked patch $x_i, \ i \in \maeT,$ based on the surrounding multimodal context. Following~\citet{MAE}, we add a shallow transformer $h_{\psi}$ as the image patch decoder on top of the VLM encoder. The MAE loss is then computed with MSE over the masked patches:
\begin{equation}
\maeLoss(\theta, \psi; \ourX) = \sum_{i \in \maeT} \ell_{\textup{MSE}}(h_{\psi}(f_{\text{bi},\theta}(\ourX))_i, x_i).
\end{equation}

The final modality-aware continual pre-training objective is a weighted sum of MLM and MAE objective on the sequence with masked texts and images, \ie, 
$$\ourLoss(\theta, \phi, \psi; \ourX) = \mlmLoss(\theta, \phi; \ourX) + w \maeLoss(\theta, \psi; \ourX),$$
where $w$ is the weight to balance these two losses\footnote{MLM and MAE loss are calculated on the same input $\ourX$, which contains masked image patches and masked text tokens.}. 

\paragraph{Efficient implementation.} In practice, we use data parallel to distribute workloads across multiple GPUs. To achieve load balancing, we calculate the compute cost of each sequence based on sequence length and image sizes and implement a sequence packing algorithm to make sure all GPUs process a batch of sequences with almost the same compute cost.

\subsection{Heterogeneous Contrastive Fine-tuning}

Following modality-aware continual pre-training, we fine-tune the bidirectional embedding model with a contrastive objective over a broad range of heterogeneous data. 
Unlike prior methods~\cite{CLIP,MMEB,mme5} that primarily rely on image-caption pairs, our approach leverages more diverse sources to improve generalization and robustness. 
As shown in the right part of Figure~\ref{fig:main2}, this includes: \textbf{(1) Long-form multimodal pairs}, which consist of document-level inputs containing both images and extended text, from VisRAG~\cite{visrag} and ViDoRe~\cite{ColPali}.
These examples support complex cross-modal reasoning and coherence over long contexts.
\textbf{(2) Curated multimodal pairs} drawn from datasets such as MMEB~\cite{MMEB} and mmE5~\cite{mme5}, which offer varied and high-quality alignments beyond typical captioning scenarios.
\textbf{(3) Text-only pairs}, sampled from large-scale retrieval datasets like E5~\cite{E5}, which enhance the model's ability to encode fine-grained semantic differences in language.

Each training instance is structured as a tuple $z=(q, d^+, \{d^-_1, \dots, d^-_K\})$, where the query $q$, the positive document $d^+$, and the hard negative documents $\{d^-_1, \dots, d^-_K\}$ may be either unimodal (text or image) or interleaved multimodal (text-image). Note that in contrastive learning, documents of other instances in the same batch, including both positive documents and negative documents, are also used as in-batch negatives, denoted by $N_{\text{in}}(z)$. Denote by $\mathbf{q}$, $\mathbf{\mathbf{d}}$ the embeddings of $q$, $d$, respectively, \ie, $\mathbf{q} = \Embbi(q)$ and $\mathbf{d} = \Embbi(d)$.
The contrastive loss is then defined as:
\begin{equation}
\mathcal{L}_{\text{CL}}(\theta;z) = -\log\frac{\Phi(\mathbf{q}, \mathbf{d}^+)}{\Phi(\mathbf{q}, \mathbf{d}^+) + \sum\limits_{k=1}^{K} \Phi(\mathbf{q}, \mathbf{d}^-_k) + \sum\limits_{d \in N_{\text{in}}(z)} \Phi(\mathbf{q}, \mathbf{d})},
\label{eq:contrastive_loss}
\end{equation}

where $\Phi(\cdot, \cdot)$ is a similarity function defined as: $
\Phi(\mathbf{a}, \mathbf{b}) = \exp\left(\text{cos}(\mathbf{a}, \mathbf{b}) / \tau\right)$,
with $\text{cos}(\cdot, \cdot)$ denoting cosine similarity and $\tau$ a temperature hyperparameter.

This objective encourages the model to assign higher similarity scores to positive pairs $(q, d^+)$ while suppressing scores for negative pairs $(q, d^-_k)$, resulting in semantically meaningful and discriminative embeddings across modalities and domains.

\paragraph{Task-aware batching.} In Heterogeneous Contrastive Fine-tuning, documents from different tasks can vary significantly. For example, a photograph of a puppy is vastly different from a screenshot of an arXiv paper. This stark difference can make distinguishing them trivial, which diminishes the benefit of in-batch negatives in the contrastive loss. To create a more effective in-batch negative sampling strategy for heterogeneous pairs, we adopt task-aware batching~\citep{GTE}. By ensuring that all instances within the same batch originate from the same task, we enable harder in-batch negative samples, leading to improved representation quality.

\section{Experiments}

\subsection{Experimental Setup}
\label{subsec:setup}

\subsubsection{Modality-aware Continual Pre-training Stage}

\noindent \textbf{Training Data.}
We incorporate three categories of training corpora as the CPT dataset:
(1) \textbf{Text-only data} from DCLM~\citep{dclm},
(2) \textbf{Common image-text pairs} from PixelProse~\citep{pixelprose} (CommonPool, CC12M, and RedCaps), MAmmoTH-VL-Instruct~\citep{mammoth}, and MMEB training set~\citep{MMEB}, and
(3) \textbf{Document-level multimodal data} from DocMatix~\citep{docmatix}, VisRAG~\citep{visrag}, and ColPali training set~\citep{ColPali}.
For each dataset, we randomly sample 500K instances, resulting in a total corpus of approximately 30B tokens.

\noindent \textbf{Implementation Details.}
We adopt Qwen-2.5-VL~\citep{Qwen25-VL} as the VLM backbone.
Training is conducted with a maximum input sequence length of 2048 tokens and a micro-batch size of 12,800 across 32 NVIDIA H100 GPUs (80GB each).
The learning rate is set to $2 \times 10^{-6}$.
For the denoising objectives, we apply a masked language modeling (MLM) probability of 0.4 for \ours{}-3B and 0.6 for \ours{}-7B.
For masked image reconstruction, we use a masking ratio of 0.5 for \ours{}-3B and 0.6 for \ours{}-7B.
The lightweight image decoder for MAE ($h_{\psi}$) is initialized from the middle layer of the Qwen-2.5-VL backbone to improve loss stability.
The MAE loss $\maeLoss$ is weighted by $w = 0.5$ to balance with the MLM loss $\mlmLoss$ in the total loss function.

\subsubsection{Heterogeneous Contrastive Learning Stage}

\noindent \textbf{Training Data.}
We use three categories of datasets for contrastive learning:
(1) \textbf{Long-form multimodal pairs} from VisRAG~\citep{visrag} and the ViDoRe training set~\citep{ColPali},
(2) \textbf{Common multimodal pairs} from the training sets of MMEB~\citep{MMEB} and mmE5~\citep{mme5}, and
(3) \textbf{Text-only pairs} from the large-scale dense retrieval dataset E5~\citep{E5}.
From each dataset, we randomly sample 50K instances, resulting in a total of approximately 2M contrastive training pairs.

\noindent \textbf{Implementation Details.} We initialize the model from the checkpoint obtained after modality-aware CPT. 
Training is performed with a batch size of 2048. We apply task-aware batching~\cite{GTE} to ensure that query-document pairs from the same dataset (task) are grouped within each batch.
The learning rate is set to $1 \times 10^{-5}$, and the temperature parameter $\tau$ in the contrastive loss is fixed at 0.03.
To improve the training signal, each positive pair is accompanied by two hard negative pairs.

\subsubsection{Evaluation}

\noindent \textbf{Massive Multimodal Embedding Benchmark (MMEB).}
We assess the general embedding quality of our model using the MMEB benchmark~\cite{MMEB}, reporting results with Precision@1.
MMEB includes 36 multimodal tasks across four types: 10 classification tasks, 10 visual question answering (VQA) tasks, 12 retrieval tasks, and 4 visual grounding tasks.

\noindent \textbf{Visual Document Retrieval Benchmarks (ViDoRe-v2).}
We evaluate our model's performance on visually rich document retrieval by NDCG@5 using the ViDoRe-v2 benchmark. 
ViDoRe~\cite{ColPali} is designed to test retrieval systems across diverse document types, tasks, languages, and settings.
The v2 version follows the BEIR~\citep{beir} evaluation format and extends ViDoRe-v1 to include multilingual and more generalized retrieval settings.

\subsection{Overall Results}

\begin{table}[ht]
\centering
\small
\caption{MMEB results, which includes 36 tasks across four categories: Classification, Visual Question Answering (VQA), Retrieval, and Visual Grounding. 
In addition to existing baselines, we evaluate three variants of Qwen-2.5-VL with different model sizes and attention mechanisms.
We highlight the best scores in \textbf{bold} and the second-best scores with an \underline{underline}.}
\begin{tabular}{lcccccccc}
\toprule
\multirow{2}{*}{Models} & \multirow{2}{*}{Size} & \multicolumn{4}{c}{Per Meta-Task Score} & \multicolumn{3}{c}{Average Score} \\
\cmidrule(r){3-6}\cmidrule(l){7-9}
 &  & Classification & VQA & Retrieval & Grounding & IND & OOD & Overall \\
\midrule
\multicolumn{9}{l}{\textit{Existing Baselines}} \\
CLIP~\citep{CLIP}                     & 428M   & 55.2 & 19.7 & 53.2 & 62.2 & 47.6 & 42.8 & 45.4 \\
BLIP2~\citep{blip2}                   & 428M   & 27.0 & 4.2  & 33.9 & 47.0 & --   & --   & 25.2 \\
OpenCLIP~\citep{OpenCLIP}             & 428M   & 56.0 & 21.9 & 55.4 & 64.1 & 50.5 & 43.1 & 47.2 \\
SigLIP~\citep{SigLIP}                 & 652M   & 40.3 & 8.4  & 31.6 & 59.5 & --   & --   & 34.8 \\
MagicLens~\citep{MagicLens}           & 613M   & 38.8 & 8.3  & 35.4 & 26.0 & --   & --   & 27.8 \\
UniIR~\citep{UniIR}                   & 428M   & 42.1 & 15.0 & 60.1 & 62.2 & --   & --   & 42.8 \\
MM-EMBED~\citep{mmembed}              & 7B   & 48.1 & 32.2 & 63.8 & 57.8 & --   & --   & 50.0 \\
GME~\citep{GME}                       & 7B   & 56.9 & 41.2 & 67.8 & 53.4 & --   & --   & 55.8 \\
VLM2Vec~\citep{MMEB}                  & 7B   & 61.2 & 49.9 & 67.4 & 86.1 & 67.5 & 57.1 & 62.9 \\
MMRet~\citep{megapairs}               & 7B   & 56.0 & 57.4 & 69.9 & 83.6 & 68.0 & 59.1 & 64.1 \\
mmE5~\citep{mme5}                     & 11B   & \textbf{67.6} & \underline{62.7} & \underline{71.0} & \underline{89.7} & \underline{72.4} & \underline{66.6} & \underline{69.8} \\
\midrule
\multicolumn{9}{l}{\textit{Variants of Qwen-2.5-VL with only Contrastive Learning on MMEB Training Set}} \\
causal attn.        & 3B  & 59.8 & 63.8 & 68.2 & 83.8 & 72.7 & 58.5 & 66.4 \\
bidirectional attn. & 3B  & 59.1 & 60.6 & 68.7 & 83.4 & 71.7 & 57.6 & 65.4 \\
bidirectional attn. & 7B  & 60.5 & 62.2 & 70.5 & 85.6 & 72.6 & 60.1 & 67.1 \\
\midrule
\multicolumn{9}{l}{\textit{Ours}} \\
\ours{}-3B & 3B & 59.8 & 62.9 & 70.6 & 88.6 & 72.3 & 61.5 & 67.5 \\
\ours{}-7B & 7B & \underline{65.8} & \textbf{64.7} & \textbf{75.0} & \textbf{92.4} & \textbf{74.7} & \textbf{67.6} & \textbf{71.5} \\
\bottomrule
\end{tabular}
\vspace{-0.1cm}
\label{tab:mmeb}
\end{table}

\begin{table*}[h]
\centering
\small
\setlength{\tabcolsep}{2.2pt}
\caption{The results on the ViDoRe-v2 benchmark, which includes 7 tasks. 
``Syn'' denotes synthetic data, ``Mul'' indicates multilingual tasks, and ``Bio'' refers to biomedical domains.
The best results are shown in \textbf{bold}, while the second-best are \underline{underlined}.}
\begin{tabular}{lccccccccc}
\toprule
Models & Size & ESG\_Human & Eco\_Mul & Bio & ESG\_Syn & ESG\_Syn\_Mul & Bio\_Mul & Eco & Avg. \\
\midrule
\multicolumn{10}{l}{\textit{Existing Baselines}} \\
SigLIP~\citep{SigLIP}        & 652M & 28.8 & 14.0 & 33.8 & 19.8 & 21.9 & 18.2 & 29.8 & 23.8 \\
VLM2Vec~\citep{MMEB}         & 7B & 33.9 & 42.0 & 38.8 & 36.7 & 38.4 & 29.7 & 51.4 & 38.7 \\
VisRAG-Ret~\citep{visrag}         & 3B & 53.7 & 48.7 & 54.8 & 45.9 & 46.4 & 47.7 & 59.6 & 51.0 \\
vdr-multi-v1\tablefootnote{\url{https://huggingface.co/llamaindex/vdr-2b-multi-v1}}         & 2B & 63.1 & 52.8 & 60.6 & 50.3 &51.2 & 56.9 & 61.2 & 56.6 \\
GME~\citep{GME}              & 7B & \textbf{65.8} & 56.2 & \textbf{64.0} & 54.3 & \textbf{56.7} & 55.1 & \underline{62.9} & \underline{59.3} \\
mmE5~\citep{mme5}            & 11B & 52.8 & 44.3 & 51.3 & 55.1 & 54.7 & 46.8 & 48.6 & 50.5 \\
\midrule
\multicolumn{10}{l}{\textit{Ours}} \\
\ours{}-3B & 3B & \underline{63.3} & \underline{57.3} & 62.5 & \textbf{58.3} & \underline{54.8} & \underline{59.8} & 62.8 & \textbf{59.8} \\
\ours{}-7B & 7B & 58.8 & \textbf{57.6} & \underline{63.2} & \underline{55.3} & 51.4 & \textbf{61.3} & \textbf{63.8} & 58.8 \\
\bottomrule
\end{tabular}
\vspace{-0.1cm}
\label{vidore-v2}
\end{table*}

We present the overall multimodal embedding performance on MMEB in Table~\ref{tab:mmeb} and the results on long-form document-level retrieval from ViDoRe-v2 in Table~\ref{vidore-v2}.
Our model, \ours{}, consistently outperforms all strong baselines on both benchmarks, demonstrating the effectiveness of our proposed framework.
Several key observations can be drawn from the results:
(1) The combination of a VLM backbone with bidirectional attention, Continual Pre-training (CPT), and heterogeneous Contrastive Learning (CL) consistently yields superior performance.
Specifically, the configuration ``bidirectional + CPT + CL'' outperforms both ``causal + CL'' and ``bidirectional + CL'' (without CPT). 
This confirms the importance of modality-aware pre-training in unlocking the full potential of bidirectional architectures.
(2) After continual pre-training on 30B tokens, our 3B model with bidirectional adaptation surpasses 7B baselines trained only with contrastive learning.
(3) Scaling our approach to a 7B model leads to substantial improvements across all MMEB task categories, establishing new state-of-the-art results on MMEB.
(4) Although the 7B model performs better on MMEB overall, the 3B model achieves slightly higher results on ViDoRe-v2. 
This is likely because ViDoRe-v2 includes fewer training and evaluation samples, and the smaller model is less prone to overfitting in such low-resource settings. Across the full MMEB benchmark, however, larger models show consistent improvements, confirming that our method scales well with model size.

\subsection{Ablation Study}\label{sec:ablation}

\begin{table*}[!t]
    \centering
    \caption{Performances of ablated models on both benchmarks. We evaluate the contribution of each component in our framework by removing key elements from the modality-aware continual pretraining stage and the heterogeneous contrastive fine-tuning stage.}
    \begin{tabular}{lcc}    
    \toprule
         Model & MMEB & Vidore-v2 \\
        \midrule
        \ours{}-3B  & \textbf{67.5}  & \textbf{59.8}  \\
        \midrule
\multicolumn{3}{l}{\textit{Modality-aware Continual Pre-training Stage}}                   \\
        \quad w/o. MLM &  66.2  &  57.2  \\
        \quad w/o. MAE &  66.8  &  56.9  \\
        \quad w/o. MLM \& MAE, \ie, no Continual Pre-training & 65.8 &  56.2  \\
        \hline
\multicolumn{3}{l}{\textit{Heterogeneous Contrastive Fine-tuning Stage}}                   \\
        \quad w/o. Text-only Pairs &  67.1 &  59.2  \\
        \quad w/o. Long-form Document Retrieval Pairs &  66.9 &  45.7  \\
        \quad w/o. Task-aware Batching &  67.2 &  58.8  \\
    \bottomrule
    \end{tabular}
\vspace{-0.1cm}
    \label{tab:ablation}
\end{table*} 

To understand the contribution of each major design choice in our framework, we conduct ablation studies on both the \textit{Modality-aware Continual Pre-training} and \textit{Heterogeneous Contrastive Fine-tuning} stages. 
As shown in Table~\ref{tab:ablation}, removing any key component leads to a consistent performance drop on both benchmarks (MMEB and Vidore-v2), demonstrating the importance of each part.

\noindent\textbf{Modality-aware Continual Pre-training.}
The ablation of either the \textit{Masked Language Modeling} (MLM) or the \textit{Masked Autoencoding} (MAE) objective results in a performance decline, indicating that both text and image reconstruction are essential for learning modality-specific representations. 
Disabling \emph{both} yields the largest decline, demonstrating the joint effectiveness between MLM and MAE in training a robust bidirectional embedding model.

\noindent \textbf{Heterogeneous Contrastive Fine-tuning.}
We then evaluate the effect of different types of training data used during contrastive fine-tuning. 
Removing \textit{text-only pairs} results in a noticeable drop, showing their importance for maintaining strong language representations. 
Excluding \textit{long-form document retrieval pairs} (VisRAG and the training set of ColPali) also hurts performance, especially on ViDoRe, demonstrating their value in supporting deeper reasoning over extended contexts. 
Finally, removing the \textit{task-aware batching} technique, where data from different tasks are jointly trained in each batch, leads to performance degradation.
This technique helps prevent the model from overfitting to task-specific patterns and encourages it to learn more sample-discriminative representations, which is crucial for generalization across diverse task formats.


\subsection{Data Scaling Effect for Continual Pre-training}

To evaluate the impact of data scale on the effectiveness of continual pre-training (CPT), we analyze how downstream performance changes with increasing CPT steps.
Specifically, we perform contrastive learning (CL) using checkpoints saved at different stages of a single CPT run for both 3B and 7B models.

\begin{figure}[t]
\centering
\includegraphics[width=0.9\textwidth]{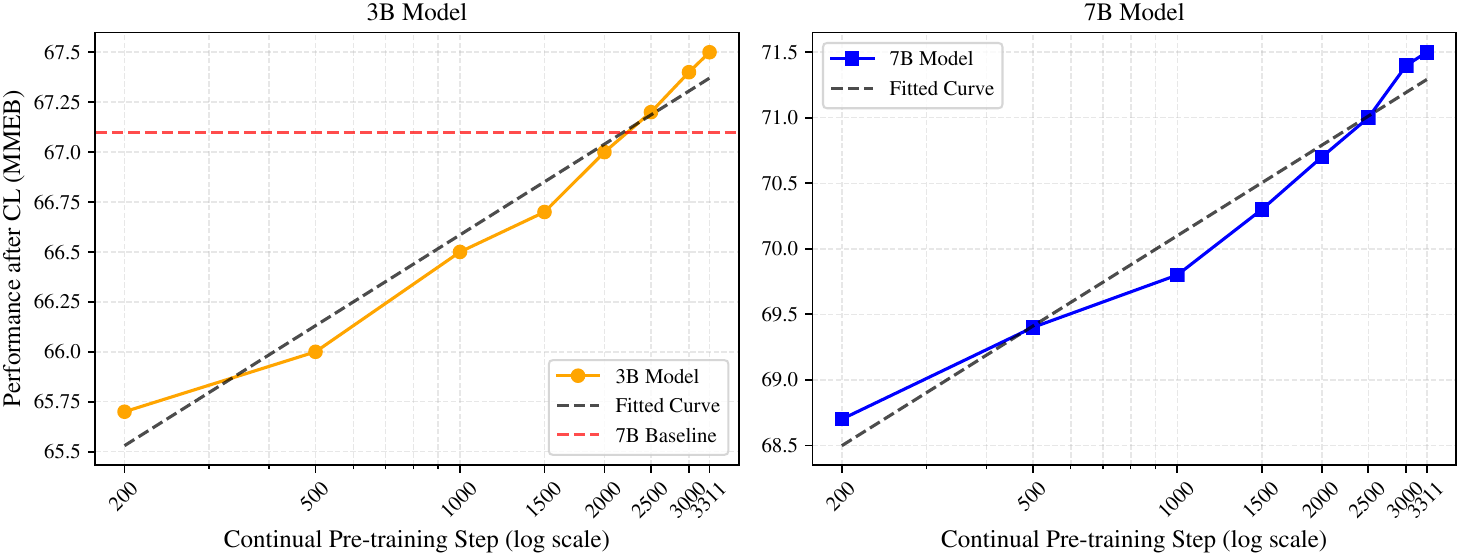}
\vspace{-2pt}
\caption{Scaling effect of our CPT stage on downstream performance. 
We evaluate MMEB performance after CL using checkpoints (left: 3B, right: 7B) from different steps of CPT.}
\label{fig:scaling_law} 
\end{figure}

As shown in Figure~\ref{fig:scaling_law}, downstream performance on MMEB improves consistently as the number of CPT steps increases.
Notably, after approximately 2,200 steps (corresponding to 20B tokens), the 3B model achieves performance on par with the 7B baseline trained without CPT.
This demonstrates that modality-aware continual pre-training substantially enhances the quality of bidirectional representations, which in turn improves alignment during the CL stage.

While constrained by computational resources, our findings suggest that further scaling of CPT with more data and training steps can continue to improve model performance.
This insight provides practical guidance for balancing training cost with expected gains in future work.

\subsection{Hyperparameter Analysis of Continual Pre-training}

\begin{figure}[t]
\centering
\includegraphics[width=0.99\textwidth]{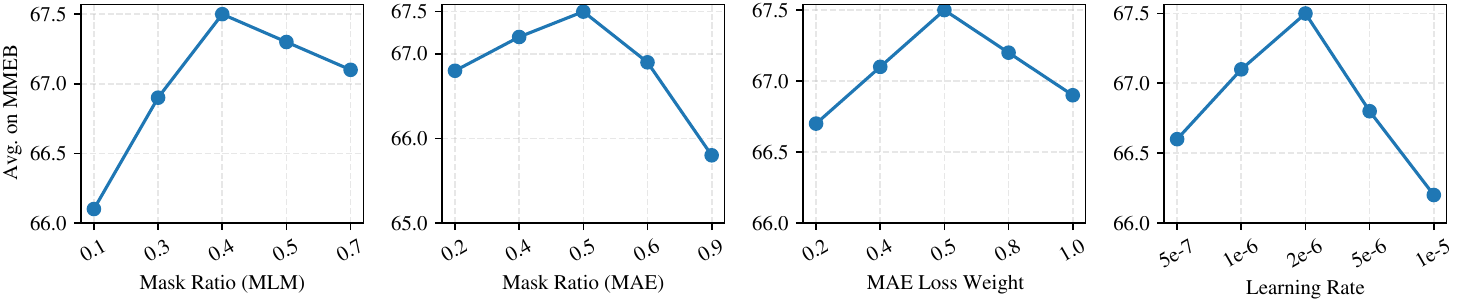}
\vspace{-2pt}
\caption{The performances of \ours{} (3B) with different CPT settings on MMEB.}
\label{fig:param} 
\end{figure}

To further understand the training process of the CPT stage, we conduct experiments of hyperparameter analysis and present the results in Figure~\ref{fig:param}. 
We evaluate the performance of \ours{} (3B) on MMEB using models trained with a fixed amount of data. 
We select hyperparameter values based on performance on validation sets, each containing 1K samples drawn from the corresponding training data. 
To maintain consistency with earlier experiments, we report the results on the MMEB test set.

\noindent \textbf{Mask Ratio.}
We examine the effect of different masking probabilities for both masked language modeling (MLM) and masked autoencoding (MAE). 
Increasing the mask ratio generally encourages the model to rely more on contextual signals across modalities, but overly high ratios can lead to degraded learning due to excessive information removal and low signal-to-noise ratio. 
Our experiments show that moderate masking rates (MLM: 40\%, MAE: 50\%) strike a good balance, enabling strong cross-modal reasoning without making the model to forget relevant input tokens or patches.

\noindent \textbf{Loss Weight.}
We also study the weight assigned to the MAE loss relative to MLM in the overall CPT objective. 
A balanced combination is crucial: if the MAE loss is underweighted, the model may neglect visual reconstruction and fail to integrate visual semantics effectively; 
if overweighted, it may distort the language modeling objective. 
We find that setting the MAE loss weight to 0.5 provides the best trade-off, aligning well with both visual and textual learning objectives.

\noindent \textbf{Learning Rate.}
The learning rate is a critical factor influencing the stability and convergence speed of continual pretraining. 
A lower learning rate can lead to underfitting, especially in early training stages, while an overly large learning rate may disrupt the pre-trained knowledge and destabilize training. 
Through empirical tuning, we find that a learning rate of $2 \times 10^{-6}$ provides a stable optimization process, allowing the model to adapt effectively to the new denoising objectives without catastrophic forgetting.

\section{Related Work}

\noindent \textbf{Multimodal Embedding.}
Multimodal embedding aims to represent inputs from different modalities in a shared space to support cross-modal understanding and interaction.
Early models in this area include ALIGN~\cite{ALIGN}, BLIP~\cite{BLIP}, and CLIP~\cite{CLIP}, which adopt dual-encoder architectures. 
These models encode each modality separately and align their outputs using contrastive learning.
Recent work builds on stronger vision-language models (VLMs). For example, VLM2Vec~\cite{MMEB} is based on Phi-3.5-V~\cite{Phi3}, GME~\cite{GME} is built on LLaVA~\cite{Llava}, and mmE5~\cite{mme5} uses mLLaMA~\cite{llama3}.
Some methods also improve the fine-tuning stage. LLaVE~\cite{LLaVE}, for instance, applies hard negative mining and task-aware batch sampling to improve alignment.
However, most of these approaches still rely on causal models and do not explore the advantages of bidirectional architectures.
We argue that, with appropriate adaptation, bidirectional VLMs can produce stronger and more robust multimodal embeddings.

\noindent \textbf{Continual Pre-training for Multimodal Models.}
Continual pre-training (CPT) involves further training of pre-trained models using additional data or new objectives to improve performance or adapt to specific downstream tasks~\cite{CPT_survey}.
In multimodal learning, models such as LXMERT~\cite{LXMERT} and UNITER~\cite{UNITER} apply Masked Language Modeling (MLM) during pre-training to jointly learn image-text representations, often using relatively small transformer architectures.
Another line of work explores CPT using reconstruction-based objectives in multimodal settings~\cite{ViLT,Janus,JanusFlow,Janus-Pro,SEED-X,QLIP,VLMo,Emu3,TiTok}.
For example, ViLT~\cite{ViLT} uses both MLM and Image-Text Matching (ITM) during training. More recently, the Janus models~\cite{Janus,JanusFlow,Janus-Pro} from DeepSeek apply reconstruction losses to both text tokens and image pixels.
Motivated by these prior efforts, we propose a CPT framework that adapts VLMs into strong bidirectional multimodal embedding models through modality-aware pre-training strategies.

\section{Conclusion}

In this paper, we have proposed a two-stage framework for multimodal embeddings, combining modality-aware continual pre-training and heterogeneous contrastive fine-tuning. 
Our method \ours{} leverages bidirectional attention mechanisms and joint reconstruction objectives to enhance cross-modal interactions. 
Additionally, by incorporating diverse and extensive multimodal data, our framework significantly improves the robustness and generalization of embedding models. 
Experimental results show that our approach achieves state-of-the-art performance, demonstrating strong scalability with respect to both model and data size on MMEB. 
This study shows the effectiveness and potential of continual pre-training strategies for advancing multimodal embedding research.

However, several directions remain for future exploration.
First, extending continual pre-training to incorporate more diverse modalities such as video, speech, or structured data could further enhance the generality of learned embeddings.
Second, investigating more advanced denoising objectives or unified encoder-decoder architectures may improve the efficiency and representation quality of bidirectional multimodal models.
Third, evaluating on more complex real-world applications, such as multi-hop retrieval or interleaved-inputs scenarios, would further validate the robustness of \ours{}.


\appendix

\clearpage

\section*{Appendix}

\section{Detailed Results on MMEB}

We present the detailed results of \ours{} and baseline models on the MMEB benchmark~\cite{MMEB} in Table~\ref{tab:detail_mmeb}, covering 36 tasks across four categories: classification, VQA, retrieval, and visual grounding. 

\begin{table*}[ht]
\centering
\caption{Detailed performance of multimodal models on 36 MMEB tasks~\cite{MMEB}. 
We show results of baseline models and our method (\ours{}) at 3B and 7B scales.}
\renewcommand{\arraystretch}{0.9}
\setlength{\tabcolsep}{2.5pt}
\resizebox{\textwidth}{!}{
\begin{tabular}{lccccccccc}
\toprule
{Task} & {CLIP} & {OpenCLIP} & {SigLIP}
 & {BLIP2} & {VLM2Vec} & {MMRet} & {mmE5} & {\ours{} (3B)} & {\ours{} (7B)} \\
\midrule
\multicolumn{10}{l}{{Classification (10 tasks)}} \\
ImageNet-1K    & 55.8 & 63.5 & 45.4 & 10.3 & 74.5 & 58.8 & 77.6 & {75.4} & {78.0} \\
N24News        & 34.7 & 38.6 & 13.9 & 36.0 & 80.3 & 71.3 & 82.1 & {80.9} & {81.5} \\
HatefulMemes   & 51.1 & 51.7 & 47.2 & 49.6 & 67.9 & 53.7 & 64.3 & {70.6} & {77.6} \\
VOC2007        & 50.7 & 52.4 & 64.3 & 52.1 & 91.5 & 85.0 & 91.0 & {87.0} & {90.0} \\
SUN397         & 43.4 & 68.8 & 39.6 & 34.5 & 75.8 & 70.0 & 77.9 & {74.8} & {76.8} \\
\rowcolor[HTML]{FFEB99} Place365       & 28.5 & 37.8 & 20.0 & 21.5 & 44.0 & 43.0 & 42.6 & {38.8} & {43.0} \\
\rowcolor[HTML]{FFEB99} ImageNet-A     & 25.5 & 14.2 & 42.6 & 3.2  & 43.6 & 36.1 & 56.7 & {39.7} & {52.7} \\
\rowcolor[HTML]{FFEB99} ImageNet-R     & 75.6 & 83.0 & 75.0 & 39.7 & 79.8 & 71.6 & 86.3 & {75.4} & {83.0} \\
\rowcolor[HTML]{FFEB99} ObjectNet      & 43.4 & 51.4 & 40.3 & 20.6 & 39.6 & 55.8 & 62.2 & {31.3} & {45.2} \\
\rowcolor[HTML]{FFEB99} Country-211    & 19.2 & 16.8 & 14.2 & 2.5  & 14.7 & 14.7 & 34.8 & {24.0} & {30.4} \\
\rowcolor[HTML]{E2F0D9}\textit{All Classification} & 42.8 & 47.8 & 40.3 & 27.0 & 61.2 & 56.0 & 67.6 & {59.8} & {65.8} \\
\midrule
\multicolumn{10}{l}{{VQA (10 tasks)}} \\
OK-VQA            & 7.5 & 11.5 & 2.4 & 8.7 & 69.0 & 73.3 & 67.9 & {40.0} & {36.9} \\
A-OKVQA           & 3.8 & 3.3  & 1.5 & 3.2 & 54.4 & 56.7 & 56.4 & {54.6} & {57.1} \\
DocVQA            & 4.0 & 5.3  & 4.2 & 2.6 & 52.0 & 78.5 & 90.3 & {93.0} & {94.3} \\
InfographicsVQA   & 4.6 & 4.6  & 2.7 & 2.0 & 30.7 & 39.3 & 56.2 & {67.7} & {77.2} \\
ChartQA           & 1.4 & 1.5  & 3.0 & 0.5 & 34.8 & 41.7 & 50.3 & {64.1} & {69.8} \\
Visual7W          & 4.0 & 2.6  & 1.2 & 1.3 & 49.8 & 49.5 & 51.9 & {61.6} & {58.5} \\
\rowcolor[HTML]{FFEB99} ScienceQA         & 9.4 & 10.2 & 7.9 & 6.8 & 42.1 & 45.2 & 55.7 & {45.4} & {59.2} \\
\rowcolor[HTML]{FFEB99} VizWiz            & 8.2 & 6.6  & 2.3 & 4.0 & 43.0 & 51.7 & 52.8 & {52.3} & {46.2} \\
\rowcolor[HTML]{FFEB99} GQA               & 41.3& 52.5 & 57.5 & 9.7 & 61.2 & 59.0 & 62.1 & {66.9} & {71.6} \\
\rowcolor[HTML]{FFEB99} TextVQA           & 7.0 & 10.9 & 1.0 & 3.3 & 62.0 & 79.0 & 83.5 & {83.1} & {75.8} \\
\rowcolor[HTML]{E2F0D9}\textit{Avg. VQA} & 9.1 & 10.9 & 8.4 & 4.2 & 49.9 & 57.4 & 62.7 & {62.9} & {64.7} \\
\midrule
\multicolumn{10}{l}{{Retrieval (12 tasks)}} \\
VisDial        & 30.7 & 25.4 & 21.5 & 18.0 & 80.9 & 83.0 & 73.7 & {80.5} & {84.5} \\
CIRR           & 12.6 & 15.4 & 15.1 & 9.8  & 49.9 & 61.4 & 54.9 & {55.7} & {53.4} \\
VisualNews\_t2i& 78.9 & 74.0 & 51.0 & 48.1 & 75.4 & 74.2 & 77.7 & {74.4} & {78.2} \\
VisualNews\_i2t& 79.6 & 78.0 & 52.4 & 13.5 & 80.0 & 78.1 & 83.4 & {77.8} & {83.1} \\
MSCOCO\_t2i    & 59.5 & 63.6 & 58.3 & 53.7 & 75.7 & 78.6 & 76.2 & {76.4} & {79.8} \\
MSCOCO\_i2t    & 57.7 & 62.1 & 55.0 & 20.3 & 73.1 & 72.4 & 73.6 & {72.6} & {73.9} \\
NIGHTS         & 60.4 & 66.1 & 62.9 & 56.5 & 65.5 & 68.3 & 68.8 & {67.4} & {66.7} \\
WebQA          & 67.5 & 62.1 & 58.1 & 55.4 & 87.6 & 90.2 & 88.1 & {90.6} & {91.4} \\
\rowcolor[HTML]{FFEB99} FashionIQ      & 11.4 & 13.8 & 20.1 & 9.3  & 16.2 & 54.9 & 28.6 & {22.2} & {28.9} \\
\rowcolor[HTML]{FFEB99} Wiki-SS-NQ     & 55.0 & 44.6 & 55.1 & 28.7 & 60.2 & 24.9 & 65.2 & {73.3} & {82.7} \\
\rowcolor[HTML]{FFEB99} OVEN           & 41.1 & 45.0 & 56.0 & 39.5 & 56.5 & 87.5 & 77.3 & {75.9} & {80.4} \\
\rowcolor[HTML]{FFEB99} EDIS           & 81.0 & 77.5 & 23.6 & 54.4 & 87.8 & 65.6 & 83.6 & {80.8} & {96.9} \\
\rowcolor[HTML]{E2F0D9}\textit{Avg. Retrieval} & 53.0 & 52.3 & 31.6 & 33.9 & 67.4 & 69.9 & 71.0 & {70.6} & {75.0} \\
\midrule
\multicolumn{10}{l}{{Visual Grounding (4 tasks)}} \\
MSCOCO               & 33.8 & 34.5 & 46.4 & 28.9 & 80.6 & 76.8 & 85.0 & {80.2} & {84.6} \\
\rowcolor[HTML]{FFEB99} RefCOCO              & 56.9 & 54.2 & 70.8 & 47.4 & 88.7 & 89.8 & 92.7 & {92.1} & {94.0} \\
\rowcolor[HTML]{FFEB99} RefCOCO-matching     & 61.3 & 68.3 & 50.8 & 59.5 & 84.0 & 90.6 & 88.9 & {92.8} & {95.5} \\
\rowcolor[HTML]{FFEB99} Visual7W-pointing    & 55.1 & 56.3 & 70.1 & 52.0 & 90.9 & 77.0 & 92.3 & {89.5} & {95.3} \\
\rowcolor[HTML]{E2F0D9}\textit{Avg. Grounding} & 51.8 & 53.3 & 59.5 & 47.0 & 86.1 & 83.6 & 89.7 & {88.7} & {92.4} \\

\midrule
\multicolumn{10}{l}{{Final Score (36 tasks)}} \\
\rowcolor[HTML]{FFF2CC} \textit{All IND Avg.} & 37.1 & 39.3 & 32.3 & 25.3 & 67.5 & 59.1 & 72.4 & {72.3} & {74.7} \\
\rowcolor[HTML]{FFF2CC} \textit{All OOD Avg.} & 38.7 & 40.2 & 38.0 & 25.1 & 57.1 & 68.0 & 66.6 & {61.5} & {67.6} \\
\rowcolor[HTML]{D9E1F2} \textit{All Tasks Avg.} & 37.8 & 39.7 & 34.8 & 25.2 & 62.9 & 64.1 & 69.8 & {67.5} & {71.5} \\

\bottomrule
\end{tabular}
}
\label{tab:detail_mmeb}
\end{table*}

\end{document}